\documentclass[conference]{IEEEtran}
\IEEEoverridecommandlockouts

\usepackage{cite}

\usepackage{amsmath,amssymb,amsfonts}

\usepackage{algorithmic}

\usepackage{graphicx}

\usepackage{textcomp}

\usepackage{xcolor}

\def\BibTeX{{\rm B\kern-.05em{\sc i\kern-.025em b}\kern-.08em

T\kern-.1667em\lower.7ex\hbox{E}\kern-.125emX}}

\begin{document}

\title{Particle Swarm and EDAs}

\author{\IEEEauthorblockN{Alison Jenkins, Vinika Gupta, Alexis Myrick, and Mary Lenoir}}

\maketitle

\begin{abstract}

The Particle Swarm Optimization (PSO) algorithm is developed for solving the Schaffer F6 function in fewer than \begin{math}4000\end{math} function evaluations on a total of \begin{math}30\end{math} runs. Four variations of the Full Model of Particle Swarm Optimization (PSO) algorithms are presented which consist of combinations of Ring and Star topologies with Synchronous and Asynchronous updates. The Full Model with combinations of Ring and Star topologies in combination with Synchronous and Asynchronous Particle Updates is explored.

\end{abstract}


\section{Introduction}

The Full Model PSO can be modeled using combinations of Ring and Star topologies in combination with Synchronous and Asynchronous Particle Updates. The four types of Particle Swarm Optimization (PSO) algorithm are the Full Model, Cognition Model, Social Model, and Selfless Model. The Full Model learns from itself and others \begin{math}\phi_{1} > 0\end{math}, \begin{math}\phi_{2} > 0\end{math}. The Cognition Model learns from itself \begin{math}\phi_{1} > 0\end{math}, \begin{math}\phi_{2} = 0\end{math}. The Social Model learns from others \begin{math}\phi_{1} = 0\end{math}, \begin{math}\phi_{2} > 0\end{math}. The Selfless Model learns from others \begin{math}\phi_{1} = 0\end{math}, \begin{math}\phi_{2} > 0\end{math}, except for the best particle in the swarm, which learns from changing itself randomly (\begin{math}g \neq i\end{math})~\cite{b4}.

There are two types of PSO topologies: Ring and Star. The star topology is dynamic, but the ring topology is not. For the star neighborhood topology, the social component of the particle velocity update reflects information obtained from all the particles in the swarm~\cite{b1}.

There are two types of particle update methods: asynchronous and synchronous. The asynchronous method updates the particles one at a time, while the synchronous method updates the particles all at ones. The asynchronous update method is similar to the Steady-State Genetic Algorithm update method, while the synchronous update method is similar to the Generational Genetic Algorithm update method. The Asynchronous Particle Update Method allows for newly discovered solutions to be used more quickly~\cite{b4}. Synchronous updates are done separately from particle position updates. Asynchronous updates calculate the new best positions after each particle position update and have the advantage of being given immediate feedback about the best regions of the search space. Feedback with synchronous updates is only given once per iteration. Carlisle and Dozier reason that asynchronous updates are more important for \textit{lbest} PSO where immediate feedback will be more beneficial in loosely connected swarms, while synchronous updates are more appropriate for \textit{gbest} PSO~\cite{b1}.

Having the algorithm terminate when a maximum number of iterations, or function evaluations, has been exceeded is useful when the objective is to evaluate the best solution found in a restricted time period~\cite{b1}.

\section{Methodology}

In PSO, the vectors are \begin{math}\textbf{x} = <x_{k0},x_{k1},...,x_{kn-1}>\end{math}, \begin{math}\textbf{p} = <p_{k0},p_{k1},...,p_{kn-1}>\end{math}, and \begin{math}\textbf{v} = <v_{k0},v_{k1},...,v_{kn-1}>\end{math}, where \begin{math}k\end{math} represents the particle and \begin{math}n\end{math} represents the dimension. The x-vector represents the current position in search space. The p-vector represents the location of the best solution found so far by the particle. The v-vector represents the gradient (direction) that the particle will travel if undisturbed~\cite{b4}.

The Fitness Values are \begin{math}x_{fitness}(i)\end{math} and \begin{math}p_{fitness}(i)\end{math}. The x-fitness records the fitness of the x-vector. The p-fitness records the fitness of the p-vector~\cite{b4}.

\subsection{Ring Topology with Synchronous Particle Update PSO}

Ring Topology with Synchronous Particle Update PSO (RS PSO) is used for sparsely connected population so as to speed up convergence. In this case the particles have predefined neighborhood based on their location in the topological space. The connection between the particles increases the convergence speed which causes the swarm to focus on the search for local optima by exploiting the information of solutions found in the neighborhood. Synchronous update provides feedback about the best region of the search space once every iteration when all the particles have moved at least once from their previous position.

\subsection{Ring Topology with Asynchronous Particle Update PSO}

The Ring Topology with Asynchronous Particle Update PSO (RA PSO) has information move at a slower rate through the social network, so convergence is slower, but larger parts of the search space are covered compared to the star structure. This provides better performance in terms of the quality of solutions found for multi-modal structures than those found using the star structure. Asynchronous updates provide immediate feedback about the best regions of the search space, while synchronous updates only provide feedback once per iteration.

\subsection{Star Topology with Synchronous Particle Update PSO}

The Star Topology with Synchronous Particle Update PSO (SS PSO) uses a global neighborhood with the star topology. Whenever searching for the best particle, it checks every particle in the swarm instead of just the neighborhood of three used in a ring topology. The synchronous update only provides feedback once each cycle, so all the particles in the swarm will update their positions before more feedback is provided, instead of checking to see if one of the recently updated particles has a better fit than the particle deemed best fit at the beginning of the cycle.

\subsection{Star Topology with Asynchronous Particle Update PSO}

The Star Topology with Asynchronous Particle Update PSO (SA PSO) has particles moving all at once in the search space, which allows for newly discovered solutions to be used more quickly. The Star Topology uses a global neighborhood, meaning that the entire swarm can communicate with one another and each particle bases its search off of the global best particle known to the swarm. The benefit of using a global neighborhood is that it allows for quicker convergence since the best known particle is communicated to all the particles in the swarm.

\section{Experiment}

The experiment consists of four instances of a Full Model PSO with a cognition learning rate, \begin{math}\phi_{1}\end{math}, and a social learning rate, \begin{math}\phi_{2}\end{math}, equal to 2.05. To regulate the velocity and improve the performance of the PSO, the constriction coefficient implements to ensure convergence.

The inertia weight, \begin{math}\omega\end{math}, is also implemented to control the exploration and exploitation abilities of the swarm. Both topologies in this experiment use an \begin{math}\omega\end{math} value of 1.0, in order to facilitate exploration and increase diversity. The particles in this experiment are updated in two different ways: synchronously, and asynchronously.

Asynchronous Particle Update is a method that updates particles one at a time and allows newly discovered solutions to be used more quickly, while Synchronous Particle Update is a method that updates all the particles at once. The four instances of the PSO are variations of the two Particle Update methods, and the two topologies described.

With these four instances of the PSO, a population of 30 particles is evolved and each particle's fitness is evaluated; this is done 30 times for each PSO. The number of function evaluations is observed after each population of 30 is evolved, and these 30 best function evaluation values for 30 runs are used to perform ANOVA tests and T-Tests to determine the equivalence classes of the four instances of the PSO.

\subsection{Ring Topology with Synchronous Particle Update PSO}

The RS PSO updates synchronously at the end of every iteration. It uses ring topology to compare and select the best solution within the neighborhood of three.

\subsection{Ring Topology with Asynchronous Particle Update PSO}

The RA PSO updates asynchronously, which allows for quick updates, and uses ring topology to compare solutions within a neighborhood of three.

\subsection{Star Topology with Synchronous Particle Update PSO}

The SS PSO updates synchronously, which only allows for one update per iteration, and uses star topology to compare solutions with a global neighborhood.

\subsection{Star Topology with Asynchronous Particle Update PSO}

The SA PSO updates asynchronously, which allows for quicker updates on newly discovered solutions. The star topology uses a global neighborhood to compare solutions, which allows for quicker convergence.

\section{Results}

\begin{table}[ht]

\caption{PSO Fitness Data Set}\label{table_PSO_Dataset}

\def\arraystretch{1.1}

\begin{center}

\begin{tabular}{|c|c|c|c|c|}

\hline


\textbf{}&\textbf{}&\textbf{}&\textbf{}&\textbf{} \\


\textbf{Run} & \textbf{\textit{RS}} & \textbf{\textit{RA}} & \textbf{\textit{SS}} & \textbf{\textit{SA}} \\

\hline

1 & 4000 & 77 & 129 & 75 \\

2 & 4000 & 71 & 57 & 72 \\

3 & 82 & 82 & 82 & 65 \\

4 & 62 & 60 & 4000 & 71 \\

5 & 4000 & 72 & 49 & 56 \\

6 & 72 & 4000 & 48 & 4000 \\

7 & 95 & 4000 & 83 & 189 \\

8 & 45 & 4000 & 4000 & 4000 \\

9 & 71 & 54 & 4000 & 4000 \\

10 & 61 & 68 & 91 & 4000 \\

11 & 4000 & 66 & 38 & 89 \\

12 & 50 & 4000 & 71 & 4000 \\

13 & 4000 & 4000 & 4000 & 4000 \\

14 & 4000 & 72 & 4000 & 4000 \\

15 & 4000 & 65 & 4000 & 4000 \\

16 & 4000 & 57 & 4000 & 146 \\

17 & 54 & 69 & 58 & 4000 \\

18 & 76 & 81 & 65 & 53 \\

19 & 58 & 77 & 47 & 4000 \\

20 & 4000 & 95 & 4000 & 4000 \\

21 & 55 & 4000 & 89 & 56 \\

22 & 90 & 65 & 51 & 4000 \\

23 & 4000 & 72 & 4000 & 4000 \\

24 & 4000 & 4000 & 4000 & 73 \\

25 & 90 & 4000 & 55 & 52 \\

26 & 55 & 4000 & 4000 & 4000 \\

27 & 4000 & 58 & 61 & 40 \\

28 & 65 & 4000 & 47 & 4000 \\

29 & 62 & 4000 & 110 & 4000 \\

30 & 68 & 4000 & 68 & 64 \\

\hline

\textbf{\textit{Average}} & 1640.3667 & 1642.0333 & 1509.9667 & 2170.0333 \\

\hline

\end{tabular}


\end{center}

\end{table}

\begin{table}[ht]

\caption{Anova Test Summary}\label{table_Anova_Single_Factor_Summary}

\def\arraystretch{1.1}

\begin{center}

\begin{tabular}{|c|c|c|c|c|}

\hline


\textbf{}&\textbf{}&\textbf{}&\textbf{}&\textbf{} \\

\textbf{Groups} & \textbf{Count} & \textbf{Sum} & \textbf{Average} & \textbf{Variance} \\

\hline

\textbf{\textit{RS}} & 30 & 49211 & 1640.3667 & 2840033.757 \\

\textbf{\textit{RA}} & 30 & 49261 & 1642.0333 & 3834547.482 \\

\textbf{\textit{SS}} & 30 & 45299 & 1509.9667 & 3713758.378 \\

\textbf{\textit{SA}} & 30 & 65101 & 2170.0333 & 3959879.413 \\

\hline

\end{tabular}


\end{center}

\end{table}

\begin{table}[ht]

\caption{Anova Test Variation Summary}\label{table_Anova_Var}

\def\arraystretch{1.1}

\begin{center}

\begin{tabular}{|c|c|c|c|c|c|c|}

\hline


\textbf{}&\textbf{}&\textbf{}&\textbf{}&\textbf{}&\textbf{}&\textbf{} \\

\textbf{\begin{tabular}{@{}c@{}}Source of\\Variation\end{tabular}}&\textbf{SS}&\textbf{df}&\textbf{MS}&\textbf{F}&\textbf{P-value}&\textbf{F crit}\\

\hline

\textbf{\textit{Between}}&7721005&3&2573668&0.7&0.6&2.7\\

\textbf{\textit{Within}}&445098352&116&3837055&&&\\

\textbf{\textit{Total}}&452819357&119&&&&\\

\hline

\end{tabular}


\end{center}

\end{table}

The results place all four algorithms in the same equivalence class using both the ANOVA and Student T-tests. When the ANOVA test and T-Test are performed, the ANOVA test of the four algorithms yields a p-value of 0.57, so the F-Test is then performed to determine which two-tailed two-sample T-Test to use. In each comparison between algorithms, the T-Test results in a t Stat value that is smaller than the t Critical value, therefore the null hypothesis is accepted. The data set used is shown in Table~\ref{table_PSO_Dataset}, while the ANOVA test results are shown in Tables~\ref{table_Anova_Single_Factor_Summary} and~\ref{table_Anova_Var}. Representative T-tests are shown in Tables~\ref{table_tTest_RS_SS} and~\ref{table_tTest_SS_SA}.

\subsection{Ring Topology with Synchronous Particle Update PSO}

\begin{table}[ht]

\caption{t-Test: Two-Sample Assuming Equal Variances}\label{table_tTest_RS_SS}

\def\arraystretch{1.1}

\begin{center}

\begin{tabular}{|c|cc|}

\hline


\textbf{}&\textbf{}&\textbf{} \\

\textbf{} & \textbf{RS} & \textbf{SS} \\

\hline

\textbf{Mean} & 1640.367 & 1509.967 \\

\hline

\textbf{Variance} & 3840033.757 & 3713758.37 \\

\hline

\textbf{Observations} & 30 & 3 \\

\hline

\textbf{Pooled Variance} & 3776896.068 & \\

\hline

\textbf{Hypothesized Mean Difference} & 0 & \\

\hline

\textbf{df} & 58 & \\

\hline

\textbf{t Stat} & 0.2599 & \\

\hline

\textbf{P(T\begin{math}{<=}\end{math}t) one-tail} & 0.3979 & \\

\hline

\textbf{t Critical one-tail} & 1.6716 & \\

\hline

\textbf{P(T\begin{math}{<=}\end{math}t) two-tail} & 0.7959 & \\

\hline

\textbf{t Critical two-tail} & 2.0017 & \\

\hline

\end{tabular}


\end{center}

\end{table}

The RS PSO results is the better compared to SA PSO as observed from the T-test. It provides comparable quality solutions to RA PSO but is slower than RA PSO as it waits for all the particles to be updated. The SS PSO outperforms RS PSO by a significant margin as it has an appreciably lower mean than RS PSO when subjected to T-test. The T-test is shown in Table~\ref{table_tTest_RS_SS}.

\subsection{Ring Topology with Asynchronous Particle Update PSO}

The RA PSO results in better quality solutions than the SA PSO, since larger parts of the search space are covered compared to the star structure. Using the RA PSO, solutions are found more quickly than when using a RS PSO. The SS PSO is relatively more slow of an algorithm and results in solutions of lesser quality.

\subsection{Star Topology with Synchronous Particle Update PSO}

The SS PSO is found to be in the same equivalence class as all the other algorithms in the experiment. However, the mean value of the SS PSO is slightly smaller than the mean values of the other three algorithms. It appears that it is able to find solutions slightly more quickly than the algorithms using the ring topology as it compares solutions using a global neighborhood allowing for quicker convergence. The T-test is shown in Table~\ref{table_tTest_SS_SA}.

\subsection{Star Topology with Asynchronous Particle Update PSO}

\begin{table}[ht]

\caption{t-Test: Two-Sample Assuming Equal Variances}

\def\arraystretch{1.1}

\begin{center}

\begin{tabular}{|c|cc|}

\hline


\textbf{}&\textbf{}&\textbf{} \\

\textbf{} & \textbf{SS} & \textbf{SA} \\

\hline

\textbf{Mean} & 1509.967 & 2170.033 \\

\hline

\textbf{Variance} & 3713758.378 & 3959879.413 \\

\hline

\textbf{Observations} & 30 & 30 \\

\hline

\textbf{Pooled Variance} & 3836818.895 & \\

\hline

\textbf{Hypothesized Mean Difference} & 0 & \\

\hline

\textbf{df} & 58 & \\

\hline

\textbf{t Stat} & -1.305 & \\

\hline

\textbf{P(T\begin{math}{<=}\end{math}t) one-tail} & 0.0985 & \\

\hline

\textbf{t Critical one-tail} & 1.6716 & \\

\hline

\textbf{P(T\begin{math}{<=}\end{math}t) two-tail} & 0.1970 & \\

\hline

\textbf{t Critical two-tail} & 2.0017 & \\

\hline

\end{tabular}

\label{table_tTest_SS_SA}

\end{center}

\end{table}

The SA PSO is found to be in the same equivalence class as all of the other algorithms in this experiment. The mean value of the SA PSO is larger than the mean values of the other three algorithms; and the F value is found to be larger than the F crit value when comparing the SA PSO to each of the other algorithms as well, so the T-Test: Two Sample Assuming Equal Variances is performed. In each comparison of the SA PSO to the other three algorithms, the T-Test results in a t Stat value that is smaller than the t Critical two-tail value, therefore the null hypothesis is accepted that the hypothesized mean difference is zero, since the t Stat value is less than the t Critical two-tail value.

\subsection{Comparison of Run Times}

A comparison of the run times for each algorithm shows that the asynchronous algorithms run more quickly than the synchronous algorithms, and the ring algorithms result in longer run times than the star algorithms. The SA PSO algorithm has an average runtime of 22.56 seconds, based on 30 runs of the algorithm. The RS PSO algorithm has an average runtime of 18.07 seconds, based on 30 runs of the algorithm. The SS PSO algorithm has an average runtime of 7.07 seconds, based on 30 runs of the algorithm. The SA PSO algorithm has an average runtime of 4.65 seconds, based on 30 runs of the algorithm.

The SA PSO algorithm has the smallest run time, while the RS PSO algorithm has the longest run time.

\section{Conclusions}

The four different types of PSO are significant in their own way and have different applications. The Ring and Star topologies determine the scope of feedback whereas the synchronous or asynchronous method choice decides the nature of feedback.

The results indicate that all four algorithms are in the same equivalence class, so there is no statistically significant difference in their performance. The T-tests indicate that the best quality solutions are provided by Star Synchronous algorithm.

The SS PSO algorithm is the quickest algorithm, while the RS PSO algorithm is the slowest. These results are as expected and show that the asynchronous algorithms are quicker than the synchronous algorithms and the star algorithms have a significantly smaller run time than that of the ring algorithms.

\section{Breakdown of the Work}

\raggedright

Alison Jenkins - RA PSO and Introduction, Methodology, (Introduction). RA PSO part in Methodology, Experiment, and Results sections of {\LaTeX} report.

Vinika Gupta - RS PSO and Methodology (Modification and Conclusion). RS PSO part in Methodology, Experiment, and Results sections. Full editing and modification of {\LaTeX} report.

Alexis Myrick - SS PSO and Result. SS PSO part in Methodology, Experiment, and Results sections of {\LaTeX} report.

Mary Lenoir - SA PSO and Experiment. SS PSO part in Methodology, Experiment, and Results sections of {\LaTeX} report.

\vspace{12pt}

\end{document}